\documentclass[]{interact}

\usepackage[T1]{fontenc}
\usepackage{epstopdf}
\usepackage{graphicx}
\usepackage[caption=false]{subfig}
\usepackage{amsmath}
\usepackage{algorithm}
\usepackage{algpseudocode}
\usepackage{xcolor}
\colorlet{RED}{red}
\usepackage{xurl}
\usepackage{hyperref}
\usepackage{orcidlink}

\usepackage[natbibapa,nodoi]{apacite}
\begin{document}

\articletype{Research Article}

\title{Do AI Agents Know When a Task Is Simple? Toward Complexity-Aware Reasoning and Execution}

\author{
\name{Junjie Yin~\orcidlink{0000-0001-7782-7274}\textsuperscript{a,b}\thanks{CONTACT Junjie Yin. Email: jyin10@vols.utk.edu} and Xinyu Feng~\orcidlink{0000-0003-4020-3219}\textsuperscript{a}}
\affil{\textsuperscript{a}University of Tennessee, Knoxville, TN 37996, USA; \textsuperscript{b}Microsoft, Redmond, WA 98052, USA}
}

\maketitle

\begin{abstract}
Large language model (LLM) agents increasingly automate multi-step engineering
and informatics workflows, yet they rarely ask how much effort a task actually
requires. They often follow a \emph{maximum-context-first} strategy---re-reading
files and dependencies they have already seen---turning a one-line edit into a
small code-base audit. We argue the missing capability is \emph{task-aware
execution-scope estimation}: judging a task's difficulty, the information it truly
needs, and the shortest reliable path \emph{before} committing budget. We
formalize \emph{minimum-sufficient execution} and the \emph{Agent Cognitive
Redundancy Ratio} (ACRR), and propose \textbf{E3} (Estimate, Execute, Expand): the
agent estimates an initial operating point, executes a minimum viable path, and
expands scope only when verification fails. On MSE-Bench---a deterministic
benchmark of $121$ edits in a capability-controlled simulator---E3 matches the
strongest baseline's $100\%$ success while cutting cost by $85\%$, tokens by
$91\%$, and inspected files by $92\%$, and further beats a \emph{strong adaptive}
retrieval baseline by $16\%$; the gains survive held-out instruction wording and
essentially every cost weighting. A companion real-model harness (LLM-Case)
corroborates the effect on a \emph{live} \textsc{gpt-4o} agent editing a real
open-source library, with every candidate patch graded by \emph{actually running
the project's real \texttt{pytest} suite} against a measured oracle: the
over-reading is milder but real, and E3 is the leanest and fastest policy at
comparable task success---its one shortfall a provider rate-limit, not a
wrong edit. We frame this as a controlled probe of execution
redundancy, not a measurement of any deployed agent, and position task-aware
execution as a step toward \emph{engineering-grounded AI} (EGAI)---agents whose
effort is anchored in the engineering reality of the task. We release the framework
and benchmark.
\\ \resizebox{25pc}{!}{\includegraphics{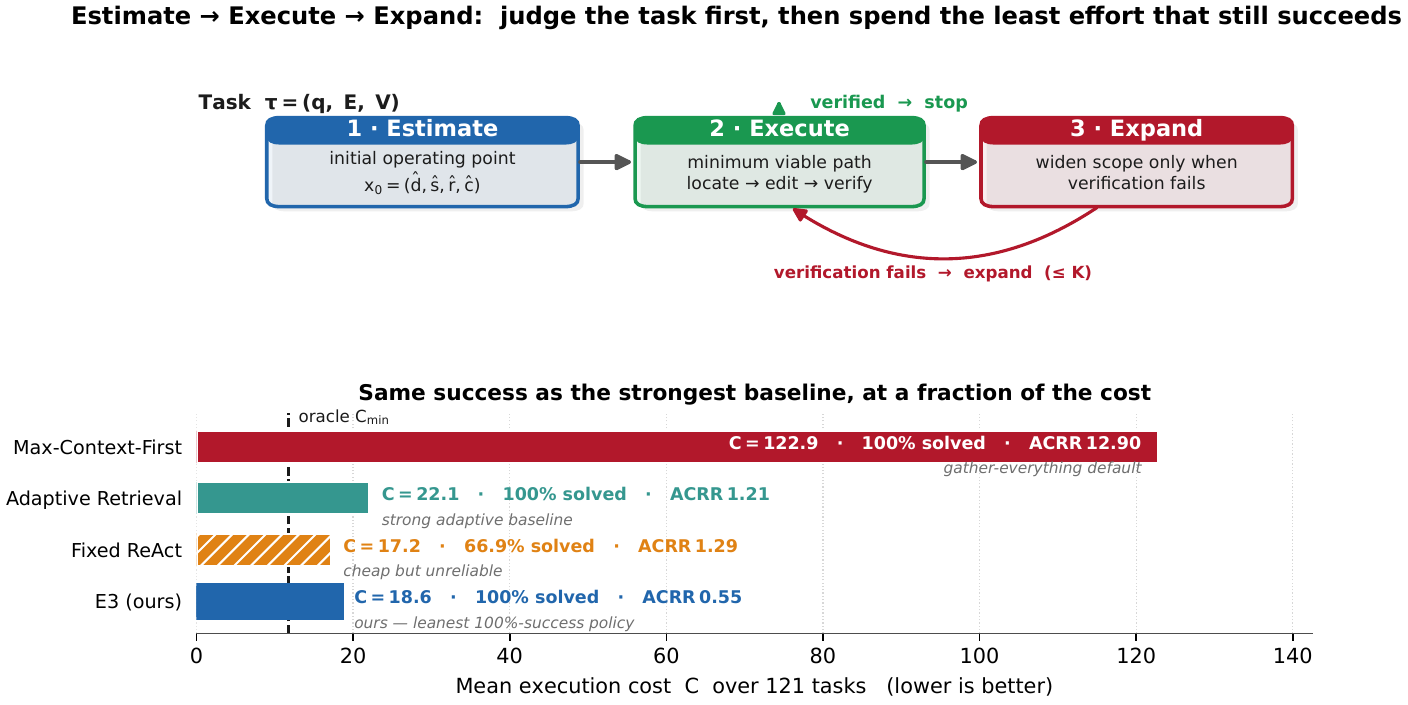}}
\end{abstract}

\begin{keywords}
LLM agents; task complexity; minimum-sufficient execution; adaptive computation; tool use; engineering informatics
\end{keywords}


\section{Introduction}\label{sec:intro}

Autonomous agents built on large language models (LLMs) now plan, call tools,
and edit artifacts across software, data, and engineering
workflows~\citep{Yao2023ReAct,Schick2023Toolformer,Xi2023AgentSurvey,Wang2024AutonomousAgentSurvey}.
Increasingly these workflows reach into open-source scientific and engineering
software---including the open-source toolchains and data-driven methods now common
in power-system analysis and control~\citep{Yang2026PVSizer,Ali2025OpenSource,Fahad2025VSG}---where
an agent's efficiency can matter as much as its correctness.
As their capability grows, so does a subtle inefficiency that is easy to feel but
hard to name. Consider a concrete, deliberately trivial task drawn from everyday
maintenance of a personal website. A home page contains two e-mail links: the
first renders a locally stored \texttt{gmail-icon.svg}; the second uses a Font
Awesome glyph, \texttt{fa-brands fa-google}. The instruction is a single
sentence: \emph{replace the second icon with the markup used by the first}. No
architectural change is requested; no external retrieval, compilation, testing,
or version-control push is needed; the required asset already exists in the
project. In essence the task is a keyword locate-and-replace.

Yet a capable frontier agent, having already edited this same project many times,
may spend several minutes on it: re-reading the icon library, re-browsing the
site directory, re-analyzing the project architecture, and re-confirming
dependencies before making a two-line change. The final edit is correct and the
reasoning never strays from the goal, but the \emph{path} to the answer is grossly
over-provisioned. A task that should take seconds is processed as a small audit.

This behavior is not a failure of knowledge or of retrieval, nor merely of
memory. It reflects a missing form of judgment: the agent cannot quickly decide
\emph{how hard the task is}, \emph{which information is necessary}, \emph{which
already-seen information is irrelevant to the current change}, and \emph{what the
shortest reliable path is}. Faced with uncertainty, many agents default to a
conservative \emph{maximum-context-first} policy: gather as much context as
possible, then eliminate every conceivable risk. On genuinely complex tasks this
caution is warranted; on simple tasks it produces large cognitive and execution
overhead.

We contend that truly efficient intelligence is not only the ability to solve
hard problems but also the ability to recognize when a problem is easy---and to
act accordingly. The aim is not to think \emph{less} for its own sake but to judge
the task \emph{correctly first}: a good estimate of what a task needs is what lets
an agent be fast and reliable at once, rather than trading caution against cost.
Human engineers do this routinely: they form a rapid estimate of
a task's difficulty and scope, sketch a minimal plan, and begin, widening their
search only if the minimal plan fails. The analogy we find most instructive comes
from power-system analysis. A power-flow solver does not enumerate the state space;
it computes a structured \emph{initial operating point}---a flat start or a DC
estimate---and then refines it with Newton--Raphson
iterations~\citep{TinneyHart1967,Stott1974LoadFlow}. A good initial point is
rarely the exact answer, but it dramatically reduces unnecessary search and makes
convergence fast and stable. Many agents today lack both the human-like judgment
of task difficulty and the solver-like initial operating point. More broadly,
judging a task's execution scope before acting---and grounding that judgment by
verification---is the agent-level expression of \emph{engineering-grounded AI}
(EGAI): anchoring an agent's reasoning and action in the physical, model-based, and
procedural reality of the task rather than in unconstrained
search~\citep{Yin2026EGAI}.

\paragraph{Research questions.} We study three questions.
(RQ1) Can we formalize the effort a task \emph{ought} to require, and measure how
far an agent's actual effort deviates from it?
(RQ2) Can an agent estimate a task's execution scope cheaply and reliably enough,
before acting, to choose a minimal execution path?
(RQ3) Does starting minimal and expanding only on failure preserve success while
cutting cost, relative to gathering maximal context up front?

\paragraph{Contributions.} We make the following contributions.
\begin{itemize}
  \item We define \emph{minimum-sufficient execution}---the least-cost agent
  trajectory that satisfies a success constraint---and the \emph{Agent Cognitive
  Redundancy Ratio} (ACRR), a normalized measure of wasted effort relative to an
  oracle (Section~\ref{sec:problem}).
  \item We propose \textbf{E3} (Estimate, Execute, Expand), a task-aware
  execution framework that constructs an initial operating point, executes a
  minimum viable path, and performs \emph{progressive context expansion} rather
  than maximum-context-first exploration (Section~\ref{sec:method}).
  \item We build \textbf{MSE-Bench}, a deterministic, offline benchmark of
  single-file, cross-file, and repository-level edits---procedurally generated
  from a small set of task archetypes and including the motivating icon
  case---in which every task carries an oracle minimum-sufficient
  trajectory, so redundancy is measurable exactly (Section~\ref{sec:bench}).
  \item We show that E3 matches the strongest baseline's success while reducing
  cost by $85\%$, tokens by $91\%$, and inspected files by $92\%$, and that it
  further improves on a \emph{strong adaptive} retrieval-augmented baseline rather
  than only an over-cautious straw man; ablations isolate the roles of estimation
  and expansion; and a power-system case study grounds the
  initial-operating-point analogy with real Newton--Raphson measurements
  (Sections~\ref{sec:results}--\ref{sec:casestudy}).
  \item We stress-test the claims: E3 retains $100\%$ success under held-out
  instruction wording that deliberately breaks the estimator's lexical cues, and
  remains the cheapest fully-successful policy under essentially every cost
  weighting---so the headline efficiency is a property of the Estimate--Expand
  \emph{architecture}, not of the estimator matching the benchmark's templates or
  of a favourable cost model (Section~\ref{subsec:robustness}).
\end{itemize}

\paragraph{Scope.} Our study is deliberately a \emph{controlled probe}. To isolate
execution redundancy from raw model capability, we evaluate policies in a
capability-invariant simulator under a stylized, configurable cost model rather
than measuring a specific deployed LLM agent. The simulator lets us construct an
exact oracle and hold ``can the agent make the edit?'' constant, so that only
\emph{how much context a policy gathers} varies. We therefore report properties of
policies \emph{under this cost model}, use hedged language throughout, and treat a
real-LLM instantiation of E3 as the primary avenue for external validation
(Section~\ref{sec:discussion}). This framing is what the phenomenon demands: the
redundancy we study is about \emph{trajectory shape}, which a capability-controlled
environment can measure cleanly and reproducibly.

\section{Related Work}\label{sec:related}

\subsection{LLM-based Autonomous Agents and Tool Use}\label{subsec:agents}
Agentic LLM systems interleave reasoning with actions such as tool calls, code
edits, and environment queries~\citep{Yao2023ReAct,Schick2023Toolformer}. They
have been deployed as open-ended explorers~\citep{Wang2023Voyager} and as software
engineers that resolve real repository issues~\citep{Jimenez2024SWEbench}. Surveys
document a rapidly expanding design space of planning, memory, and tool
interfaces~\citep{Xi2023AgentSurvey,Wang2024AutonomousAgentSurvey}. The picture
that unites these systems---an agent that plans, acts, and revises---is a
long-standing concern of this journal, from early conversational planning
agents~\citep{AllenSchubert1995TRAINS} and the integration of planning with
learning~\citep{Veloso1995PRODIGY} to software architectures for intelligent
monitoring agents~\citep{HayesRoth1996Monitoring} and, more recently, systematic
accounts of the agent design space and of multi-agent
behaviour~\citep{GrislinLeStrugeon2022AgentMining,ZhaoHernandezOrallo2025Sociality}.
Our concern, however, is orthogonal to \emph{what} tools an agent has: we study
\emph{how much} of the available context and tooling a task actually warrants.

\subsection{Reasoning and Planning Strategies}\label{subsec:reasoning}
Chain-of-thought and its relatives elicit intermediate reasoning that improves
accuracy on hard problems~\citep{Wei2022CoT,Kojima2022ZeroShot}. Self-consistency
aggregates multiple chains~\citep{Wang2023SelfConsistency}; tree- and graph-search
methods deliberate over alternatives~\citep{Yao2023ToT}; and reflection or
self-refinement revises earlier attempts~\citep{Shinn2023Reflexion,Madaan2023SelfRefine}.
Process-level verification further improves multi-step
reliability~\citep{Lightman2023VerifyStepByStep}. These methods make reasoning
\emph{deeper}; they generally assume more deliberation is beneficial and do not
ask whether a given task needs it at all. Recent analyses of ``overthinking'' in
reasoning models show that excess deliberation can waste tokens without improving,
or even while degrading, outcomes~\citep{Chen2024Overthinking}. This asymmetry
motivates our position that the \emph{difficulty appraisal} preceding reasoning
should be a first-class step rather than an afterthought---a view with deep roots
in this venue's work on metareasoning, bounded rationality, and cost-guided
planning, which we survey in
Sections~\ref{subsec:metareasoning}--\ref{subsec:planning}.

\subsection{Adaptive Computation, Routing, and Effort Allocation}\label{subsec:complexity}
A complementary line allocates compute adaptively. Classic adaptive-computation
mechanisms learn to halt~\citep{Graves2016ACT}. Test-time scaling studies how to
spend inference compute for the best accuracy-per-FLOP~\citep{Snell2024ScalingTestTime,Muennighoff2025S1}.
Cost-aware cascades and routers choose a cheaper or costlier model per
query~\citep{Chen2023FrugalGPT,Ong2024RouteLLM}. Closest to our motivation are
very recent agent-level methods: \emph{Ares} selects a per-step reasoning-effort
level and reports up to $52.7\%$ fewer reasoning tokens at similar
success~\citep{Yang2026Ares}; \emph{BoundaryRouter} decides whether a query can be
answered by direct LLM inference or must escalate to full agent
execution~\citep{Wang2026BoundaryRouter}; and \emph{Select-then-Solve} routes each
task to the best reasoning paradigm, showing that no single paradigm
dominates~\citep{Zhou2026SelectThenSolve}. These works largely perform
\emph{routing}: choosing a model, an effort level, or a paradigm from a fixed
menu. Our emphasis is different and, we argue, more basic. Rather than dialling
\emph{how much} to think or selecting \emph{which} engine to run, E3 predicts
\emph{what the task actually requires}---a structured estimate of its execution
scope, the minimum-sufficient set of files, tools, and steps---and uses that
estimate to \emph{construct} a minimum-viable plan and a good initial operating
point, widening scope only when verification demands it. We call this
\emph{execution-scope estimation} (equivalently, \emph{task-shape prediction}): an
anticipatory judgment of the right way to handle a task, not a reactive adjustment
of effort. Two properties separate it from routing (Table~\ref{tab:paradigms}).
First, the decision variable is a \emph{structured scope} that parameterises the
plan, not a scalar effort dial or a choice from a fixed set. Second, because the
estimate is deliberately optimistic, E3 pairs it with \emph{verified progressive
expansion}, so a misjudged task is recovered rather than mis-served---the judgment
exists to make the agent \emph{fast and correct at once}, not merely cheaper.
Routing chooses \emph{how much to think}; we ask \emph{what to understand before
thinking and acting}, and \emph{how little} suffices. Empirically we do not
contrast E3 only with an over-cautious default: we implement a strong
\emph{adaptive} retrieval-augmented baseline in the spirit of these routers
(Section~\ref{sec:setup}) and find that E3 still reduces cost at equal success,
isolating the value of up-front scope estimation from that of adaptivity in
general.

\begin{table}
\tbl{Where E3 sits relative to adaptive computation and routing. E3's decision
variable is a \emph{structured task scope} predicted \emph{before} acting; because
the estimate is treated as revisable, misjudgement is recovered by \emph{verified
progressive expansion} rather than paid for in a wrong or wasteful trajectory.}
{\small\begin{tabular}{@{}p{2.5cm}p{4.2cm}p{1.8cm}p{2.5cm}@{}} \toprule
Paradigm & Decision variable (what is predicted) & Timing & Recovery if misjudged \\ \midrule
Adaptive computation & \emph{how much} compute: a scalar dial (halt, depth, tokens) & during acting & --- (monotone dial) \\
Routing & \emph{which} option from a fixed menu: model, effort level, or paradigm & before acting & one-shot choice \\
Execution-scope estimation (E3, ours) & \emph{what scope} the task needs: files, tools, and steps that size a minimum-viable plan & before acting & \emph{yes}: verified progressive expansion \\ \bottomrule
\end{tabular}}
\label{tab:paradigms}
\end{table}

\subsection{Metareasoning, Introspection, and Knowing When to Think}\label{subsec:metareasoning}
If the strategies above decide \emph{how} to reason, a longer-running tradition in
this journal asks \emph{whether}---and \emph{how hard}---to reason at all.
Meta-case-based reasoning frames an agent's self-improvement as a product of
\emph{self-understanding}: reasoning about its own reasoning in order to decide how
to adapt~\citep{MurdockGoel2008MetaCBR}. Introspective reasoning monitors
performance and refines a base process only once that process proves
inadequate~\citep{FoxLeake2001Introspective,Leake1996Introspection}, and
human-like reasoning by \emph{projection} treats such anticipatory judgement as a
core cognitive mechanism~\citep{Guerin2023Projection}. These are early
formulations of the capability we argue current LLM agents lack: a cheap
meta-level judgement of how hard a task is and which already-seen information is
irrelevant to the change at hand. E3's estimation stage is a lightweight,
deliberately imperfect instantiation of exactly this introspective monitoring, and
its expansion stage is the recovery mechanism that self-monitoring accounts also
require. Because the estimator conditions on prior experience, it further connects
to case-based accounts of reusing solved episodes instead of recomputing
them~\citep{Bannour2023CBRSurvey}, while its explicit confidence output has a
natural counterpart in computational models of
trustworthiness~\citep{Primiero2025Trustworthiness}. That a cheap---even
biased---meta-level judgement can be provably worthwhile is argued from the theory
side, where regret-based meta-induction enjoys \emph{a priori} advantages that
escape the No-Free-Lunch theorem~\citep{SchurzThorn2024MetaInduction}, and from the
design side, where deliberately included cognitive biases (fast, frugal
heuristics) and ``engineered wisdom'' are held to enhance rather than degrade AI
systems~\citep{HagendorffFabi2024BiasedAI,KarlanAllen2024Wisdom}. This is our
stance in miniature: the estimator should be optimistic and inexpensive, with
verified expansion as the safety net that makes such optimism safe.

\subsection{Bounded Rationality and the Cost of Deliberation}\label{subsec:bounded}
The premise that a rational agent must economise on its own computation is the
theoretical backbone of our redundancy measure. The \emph{doubly-bounded
rationality} of an artificial agent~\citep{Sotnik2020BoundedRationality} is
precisely the lens under which the maximum-context-first policy reads as
\emph{ir}rational rather than merely cautious: an agent that eliminates every
conceivable risk on a trivial task has ignored the cost of its own deliberation.
ACRR turns this intuition into a measurement by normalising realised cost against
an oracle-defined minimum, so that ``spending five times the necessary effort''
becomes a quantity rather than a complaint; models of human decision-making under
prospect theory offer a complementary, human-factors grounding for our recurring
analogy to how skilled engineers gauge difficulty before
acting~\citep{Gupta2024HumanDecision}. Economising, however, is not the same as
always doing less. Some tasks are genuinely hard---medical diagnosis and
treatment, for instance, is shown to be NP-complete~\citep{ArleCarlson2021NPComplete}---which
is why E3 must degrade gracefully toward an exhaustive strategy in the worst case
while staying lean in the common one. The algorithmic embodiment of ``compute only
as much as the task needs'' is well established in this venue: anytime methods
return progressively better solutions under interruption~\citep{Challa2022Anytime},
and bounded solving trades exactness for tractability under an explicit
budget~\citep{Lu2024BoundedSolving}. Both share the shape of our
objective---minimise cost \emph{subject to} a reliability target, rather than
maximise certainty unconditionally.

\subsection{Planning Costs, Structure, and the Initial Operating Point}\label{subsec:planning}
Our \emph{estimate-then-execute} design has a direct antecedent in symbolic
planning. Plan-cost predictions have been used to steer domain-independent
heuristic search~\citep{Percassi2023PlanCost}: estimating the cost of completing a
plan in order to guide it is the classical-planning analogue of predicting
execution scope before committing budget. The interaction between representations
and planning objectives in decision-theoretic planning~\citep{KoenigLiu2002DTP}
prefigures our observation that the cost weights $(\alpha,\beta,\gamma,\delta)$
determine which trajectory is optimal, while hierarchical and operator-based
planning~\citep{Estlin2001HTN} and bidirectional planning that shortens search
from a structured guess~\citep{FinkBlythe2005Prodigy} mirror E3's graduated,
scope-sized execution. The complementary move---to \emph{escalate scope only on
evidence}---is the essence of goal-driven autonomy, in which an agent forms
expectations, detects discrepancies against them, and only then
re-plans~\citep{Dannenhauer2021GDA}; closely related work on planning under
temporal constraints and uncertainty~\citep{Baki2006Centralized}, on planning and
acting in dynamic environments~\citep{Chrpa2022PlanningActing}, on evaluating
plan-recovery strategies after failure~\citep{MoreiraRalha2023PlanRecovery}, and
on belief revision when new evidence arrives~\citep{Ray2018BeliefRevision}
describes, qualitatively, the monitor-and-recover loop that E3's expansion stage
makes concrete for tool-using LLM agents. Finally, the principle of retaining only
what is necessary---pruning provably redundant options through substitutability
hierarchies~\citep{WallaceFreuder2025Substitutability} or structure-exploiting
encodings~\citep{Siddiqi2024SAT}---is the search-space counterpart of the
\emph{context} pruning we advocate, and casting analogy itself as a search
procedure over a conceptual space~\citep{OstaVelezGardenfors2024Analogy}
legitimises our importing of the power-flow initial-operating-point analogy to
structure agent behaviour.

\paragraph{Summary and gap.} Taken together, this venue's literature supplies each
ingredient of our approach in isolation: agents that monitor expectations and
re-plan on failure~\citep{Dannenhauer2021GDA}; planners that estimate cost to steer
search~\citep{Percassi2023PlanCost}; introspective reasoners that judge when their
base process is inadequate~\citep{MurdockGoel2008MetaCBR,FoxLeake2001Introspective};
bounded-rational agents that must economise
computation~\citep{Sotnik2020BoundedRationality}; and anytime or bounded solvers
that trade cost for sufficiency~\citep{Challa2022Anytime,Lu2024BoundedSolving}.
What has not been provided is their \emph{unification for tool-using LLM agents},
together with an \emph{exact, capability-invariant measurement of wasted effort}.
The remainder of the paper supplies both: a formal notion of minimum-sufficient
execution and the ACRR (Section~\ref{sec:problem}), the E3 framework that
constructs and progressively revises an initial operating point
(Section~\ref{sec:method}), and MSE-Bench, whose per-task oracle makes redundancy
measurable exactly (Section~\ref{sec:bench}).

\section{Problem Formulation}\label{sec:problem}

\subsection{Agents, tasks, and trajectories}
An agent solves a task $\tau=(q,E,V)$ with natural-language query $q$,
environment $E$ (here, a repository), and an acceptance check $V$ that returns
success or failure. Acting, the agent produces a \emph{trajectory}
$\pi=(a_1,a_2,\dots,a_T)$ of tool calls $a_t\in\mathcal{A}$ (list, search, read,
inspect, trace, edit, reason, verify). Each action incurs a cost vector, and the
trajectory cost is
\begin{equation}\label{eq:cost}
C(\pi) \;=\; \alpha\,T_{\text{lat}} \;+\; \beta\,N_{\text{tok}} \;+\; \gamma\,N_{\text{tool}} \;+\; \delta\,N_{\text{file}},
\end{equation}
where $T_{\text{lat}}$ is wall-clock latency, $N_{\text{tok}}$ the reasoning and
context tokens, $N_{\text{tool}}$ the number of tool calls, and $N_{\text{file}}$
the number of distinct files pulled fully into context. The weights
$(\alpha,\beta,\gamma,\delta)$ render these axes commensurable; we report the raw
axes as well so that conclusions do not hinge on a single weighting.
Pulling an irrelevant file into context is the canonical unit of redundancy, so
$\delta$ is the largest default weight.

\subsection{Minimum-sufficient execution}
Let $P(\text{success}\mid \pi,\tau)$ denote the probability that trajectory $\pi$
passes $V$. We define the minimum-sufficient trajectory as the cheapest one that
meets a reliability target $1-\epsilon$:
\begin{equation}\label{eq:mse}
\pi^\star \;=\; \arg\min_{\pi}\; C(\pi)\quad \text{s.t.}\quad P(\text{success}\mid \pi,\tau)\;\ge\;1-\epsilon .
\end{equation}
Its cost $C_{\min}(\tau)=C(\pi^\star)$ is the effort the task \emph{ought} to
require. In a controlled environment we can construct $\pi^\star$ explicitly (an
oracle), which makes $C_{\min}$ measurable rather than hypothetical.

\subsection{Agent Cognitive Redundancy Ratio}
Given an agent's realized cost $C_{\text{act}}(\tau)$, we measure wasted effort by
the normalized excess over the minimum-sufficient cost:
\begin{equation}\label{eq:acrr}
\mathrm{ACRR}(\tau) \;=\; \frac{C_{\text{act}}(\tau) - C_{\min}(\tau)}{C_{\min}(\tau)} .
\end{equation}
$\mathrm{ACRR}=0$ denotes an optimally lean agent; $\mathrm{ACRR}=4$ means the
agent spent five times the necessary cost ($400\%$ redundancy). ACRR is defined
only for successful runs; failures are reported separately as success rate, since
a cheap failure is not an efficiency. We make no claim that the ratio itself is
novel---it is a deliberately simple, task-normalized diagnostic, excess cost over
the oracle expressed as a fraction of the minimum. Its value lies in the
normalizer: because the oracle makes $C_{\min}$ exact and \emph{per task},
redundancy becomes comparable across tasks of very different absolute cost, which
is what lets us state the paper's signature result---that redundancy is largest on
the simplest tasks.

\subsection{The initial operating point}\label{subsec:iop}
We posit that an agent should first estimate a task state
\begin{equation}\label{eq:x0}
x_0 \;=\; f(q, E, M) \;=\; (\hat d,\ \hat s,\ \hat r,\ \hat c),
\end{equation}
from the query $q$, cheap environment probes on $E$, and prior experience $M$,
where $\hat d$ is estimated difficulty, $\hat s$ the estimated scope (files or
sites to touch), $\hat r$ the risk, and $\hat c$ the confidence. The state $x_0$
is an \emph{initial operating point}: analogous to a flat start or DC estimate in
power-flow analysis~\citep{TinneyHart1967}, it need not be the answer, but a good
one keeps the subsequent search short and stable (Section~\ref{sec:casestudy}).
The agent then seeks a low-cost trajectory whose success probability meets the
target from $x_0$, expanding scope only if that fails.

\section{The E3 Framework: Estimate, Execute, Expand}\label{sec:method}

\subsection{Overview}
\begin{figure}
\centering
\includegraphics[width=0.9\linewidth]{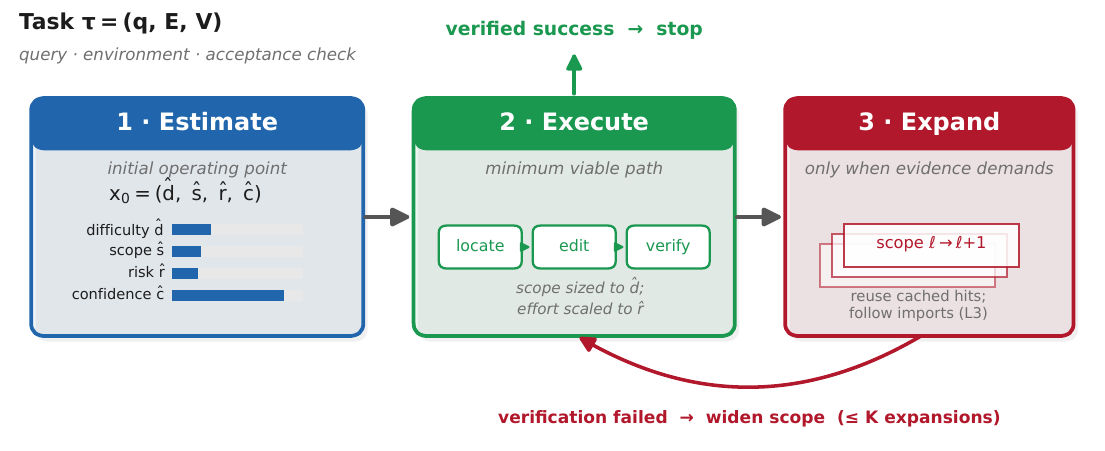}
\caption{The E3 framework. The agent estimates an initial operating point
$x_0=(\hat d,\hat s,\hat r,\hat c)$, executes a minimum viable path sized to
$x_0$, and expands scope only when verification fails or confidence is low.
Successful verification exits immediately.}
\label{fig:e3}
\end{figure}

E3 replaces the maximum-context-first loop
(\textsc{Task}$\rightarrow$\textsc{Reason}$\rightarrow$\textsc{Search}$\rightarrow$\textsc{Tool}$\rightarrow\cdots$)
with a three-stage process
(\textsc{Estimate}$\rightarrow$\textsc{Execute}$\rightarrow$\textsc{Expand}),
illustrated in Figure~\ref{fig:e3}. The agent estimates $x_0$, executes the minimum
viable path sized to $x_0$, and---only if verification fails or confidence is
low---expands scope and replans. Algorithm~\ref{alg:e3} gives the procedure.

\begin{algorithm}[tb]
\caption{E3: Estimate, Execute, Expand}\label{alg:e3}
\begin{algorithmic}[1]
\Require task $\tau=(q,E,V)$, estimator $f$, max expansions $K$
\State $x_0 \gets f(q,E,M)$ \Comment{Estimate: $\le 1$ cheap probe}
\State $\ell \gets \hat d$; \; $\mathcal{H}\gets$ cached search hits
\State $\textit{ok} \gets \Call{Execute}{\tau, \ell, x_0, \mathcal{H}}$
\State $k \gets 0$
\While{$\lnot\,\textit{ok}$ \textbf{and} $\ell < 3$ \textbf{and} $k < K$}
  \State $\ell \gets \ell + 1$; \; $k \gets k+1$ \Comment{Expand one scope level}
  \State $\textit{ok} \gets \Call{Execute}{\tau, \ell, x_0, \mathcal{H}}$
\EndWhile
\State \Return $\textit{ok}$
\end{algorithmic}
\end{algorithm}

\subsection{Estimate: task-state estimation}\label{subsec:estimate}
The estimator maps $q$ (and, when needed, one cheap probe of $E$) to $x_0$. It
combines lexical cues with an optional structural probe:
\emph{(i)} explicit file references and quoted literals with localized verbs
(``replace \dots\ in \texttt{index.html}'') signal a single-file edit;
\emph{(ii)} broad-scope cues (``refactor across the codebase'', ``every call
site'', ``re-export'') signal a repository-level change;
\emph{(iii)} otherwise the estimator spends a single \texttt{search} for the
salient token and counts occurrences to distinguish local from cross-file work.
When wording and structure conflict---localized phrasing but multiple
occurrences---confidence $\hat c$ is lowered, flagging a candidate for expansion.
Crucially, the estimator is deliberately \emph{imperfect}: some tasks that read
as local hide an indirect dependency, and will be under-estimated. The Expand
stage exists precisely to recover these cases, so the estimator can be cheap and
optimistic rather than exhaustive.

\subsection{Execute: minimum viable execution path}\label{subsec:execute}
Given $x_0$, E3 runs the smallest path likely to satisfy $V$. At scope level~1 it
localizes the single site and edits it; at level~2 it reuses the cached search
hits, reads them, and edits the direct sites; at level~3 it additionally follows
imports (\texttt{dependency\_trace}) and inspects the importer files to reach
\emph{indirect} sites that a grep cannot see. Verification effort scales with
risk: low-risk edits are checked locally, whereas high-risk repository changes run
the heavier acceptance check. The path never gathers context beyond what the
current scope level requires.

\subsection{Expand: progressive context expansion}\label{subsec:expand}
If verification fails, E3 does not restart from scratch and it does not jump to
reading everything. It increases the scope level by one, reusing what it already
learned (the remembered search hits), and replans. This \emph{progressive context
expansion} is the opposite of maximum-context-first: cost grows only in response
to evidence that the task is harder than estimated. Because expansion is bounded
by $K$ and monotone in scope, E3 degrades gracefully toward the exhaustive
strategy in the worst case while remaining lean in the common case.

\subsection{Instantiation and cost model}
Our reference implementation instantiates $\mathcal{A}$ as
\{\texttt{list\_dir}, \texttt{search}, \texttt{read\_range}, \texttt{inspect\_file},
\texttt{dependency\_trace}, \texttt{edit}, \texttt{reason}, \texttt{verify}\},
each charged per Eq.~\eqref{eq:cost}. Default weights are $\alpha{=}1.0$ per
second, $\beta{=}0.02$ per token, $\gamma{=}0.5$ per tool call, and
$\delta{=}1.5$ per fully inspected file; a full-file read also charges tokens
proportional to its length, so reading a large irrelevant file is expensive on
three axes at once. These weights are configurable, and we report a sensitivity
check in Section~\ref{subsec:sensitivity}.

\section{MSE-Bench}\label{sec:bench}

\subsection{Design principle: capability-invariant evaluation}
To isolate \emph{execution redundancy} from \emph{model capability}, MSE-Bench
uses a deterministic, offline environment in which every policy is equally able to
perform the underlying edit once it has seen the relevant content. Editability is
derived purely from what a policy has observed: a \emph{direct} site (whose
location contains the searched token literally) becomes editable once its file is
observed via search or read; an \emph{indirect} site (an alias or re-export that
the token does not match) becomes editable only after its file is fully inspected,
which in turn is discovered by dependency tracing. Consequently, policies differ
\emph{only} in how much context they gather and how they allocate effort---exactly
the phenomenon under study---and not in whether they \emph{can} make the change.
This design also makes results fully reproducible without any model API.

\paragraph{What this design does and does not claim.} Two clarifications guard
against over-reading. First, the numbers below are properties of \emph{policies
under a cost model}, not measurements of any particular commercial agent; we use
them to compare \emph{trajectory shapes}, and we hedge accordingly. Second,
holding capability fixed does \emph{not} make success trivial: reaching an
indirect site still requires the right trajectory (a dependency trace), and indeed
the Fixed ReAct policy---given \emph{identical} edit capability---fails a third of
all tasks because its fixed trajectory never reaches those sites
(Section~\ref{subsec:main}). What the design buys is a clean separation:
differences in success and cost reflect \emph{what a policy chose to look at},
not a lucky or unlucky model sample. We return to how a real, capability-varying
LLM would interact with this picture in Section~\ref{sec:discussion}.

\subsection{Task tiers and the motivating case}
MSE-Bench comprises $121$ tasks across three complexity tiers
(Table~\ref{tab:bench}). Level~1 tasks are single-file keyword replacements,
including the motivating Gmail-icon case. Level~2 tasks rename a symbol that
appears literally in a few files. Level~3 tasks are repository-level refactors
with several direct sites \emph{plus} one indirect site behind an import alias;
they come in an \emph{obvious} variant (broad-scope wording that advertises its
scope) and a \emph{deceptive} variant (localized wording that hides the indirect
dependency). Every task carries an oracle that realizes $\pi^\star$, defining
$C_{\min}$ exactly. The tasks are \emph{procedurally generated} from a small
number of archetypes---one per tier, with the Level-3 archetype split into
obvious and deceptive wordings---parameterized by randomized identifiers and
surrounded by randomized distractor files that give each repository realistic
surface area. We are explicit that this yields \emph{controlled variation} rather
than the open-ended diversity of a capability benchmark: its purpose is to make
$C_{\min}$, and hence redundancy, exactly measurable, and to admit a
\emph{held-out} paraphrased wording distribution used to stress-test the estimator
(Section~\ref{subsec:robustness}).

\begin{table}
\tbl{MSE-Bench composition. Each task ships an oracle minimum-sufficient
trajectory that defines $C_{\min}$. ``Ind.'' denotes indirect (alias/re-export)
sites reachable only by dependency tracing.}
{\begin{tabular}{@{}lcccl@{}} \toprule
Level & \#Tasks & Direct & Ind. & Oracle actions \\ \midrule
1 (local)      & 41 & 1 & 0 & locate, edit, verify \\
2 (cross-file) & 40 & 2 & 0 & search, edit${\times}2$, verify \\
3 (repo-level) & 40 & 2 & 1 & search, trace, edit${\times}3$, test \\ \midrule
\multicolumn{5}{@{}l}{\small Level 3 includes 18 deceptive tasks; the rest are obvious.} \\ \bottomrule
\end{tabular}}
\label{tab:bench}
\end{table}

\subsection{Metrics}
We report success rate, the four cost axes of Eq.~\eqref{eq:cost}, the scalar cost
$C$, and $\mathrm{ACRR}$ (Eq.~\ref{eq:acrr}). We also report the reduction of each
axis relative to the strongest baseline. Unless noted, aggregates are means over
all $121$ tasks, computed from the oracle-defined $C_{\min}$.

\section{Experimental Setup}\label{sec:setup}

\subsection{Policies}
We compare four policies plus the oracle.
\textbf{Max-Context-First (MCF)} is an \emph{upper-bound stress model} of the
over-cautious default: it walks the directory tree, fully reads every file,
reasons about the architecture, then edits and runs the heavy acceptance check. We
do \emph{not} claim that any deployed frontier agent literally reads an entire
repository; MCF bounds the cost of the ``gather everything, then act'' instinct so
that the redundancy axis is visible at full scale, and every reduction we report
is stated against a leaner baseline as well. \textbf{Fixed ReAct} is a fixed
search$\rightarrow$read-hits$\rightarrow$edit$\rightarrow$test loop with no scope
adaptation~\citep{Yao2023ReAct}. \textbf{Adaptive Retrieval (AR)} is a
\emph{strong, non-straw-man} baseline in the spirit of cost-aware routers and
retrieval-augmented agents~\citep{Ong2024RouteLLM,Yang2026Ares,Wang2026BoundaryRouter}:
it searches for the salient token, fully reads the \emph{retrieved} files, follows
their imports whenever the retrieval reveals a cross-file footprint (so indirect
sites are always reached), then edits and runs the heavy acceptance check. It is
genuinely adaptive---effort scales with the retrieved footprint---and it solves
every task; what it lacks is an up-front difficulty estimate, so it over-provisions
full reads and heavy verification on trivially local edits. The gap between AR and
E3 therefore isolates the value of \emph{scope estimation} rather than of
adaptivity in general. \textbf{E3} is our method. The \textbf{Oracle} executes
$\pi^\star$ and defines $C_{\min}$. All policies share the identical edit
capability described above.

\subsection{Implementation}
The environment, policies, benchmark, and metrics are implemented in Python with a
fixed random seed ($20260712$); all numbers below are therefore reproducible. The
estimator is a transparent lexical-plus-one-probe rule (Section~\ref{subsec:estimate})
so that the benchmark requires no external model; the same interface accepts an
LLM-backed estimator as a drop-in. We use the default cost weights unless stated.

\section{Results and Analysis}\label{sec:results}

\subsection{Main results}\label{subsec:main}
\begin{figure}
\centering
\includegraphics[width=\linewidth]{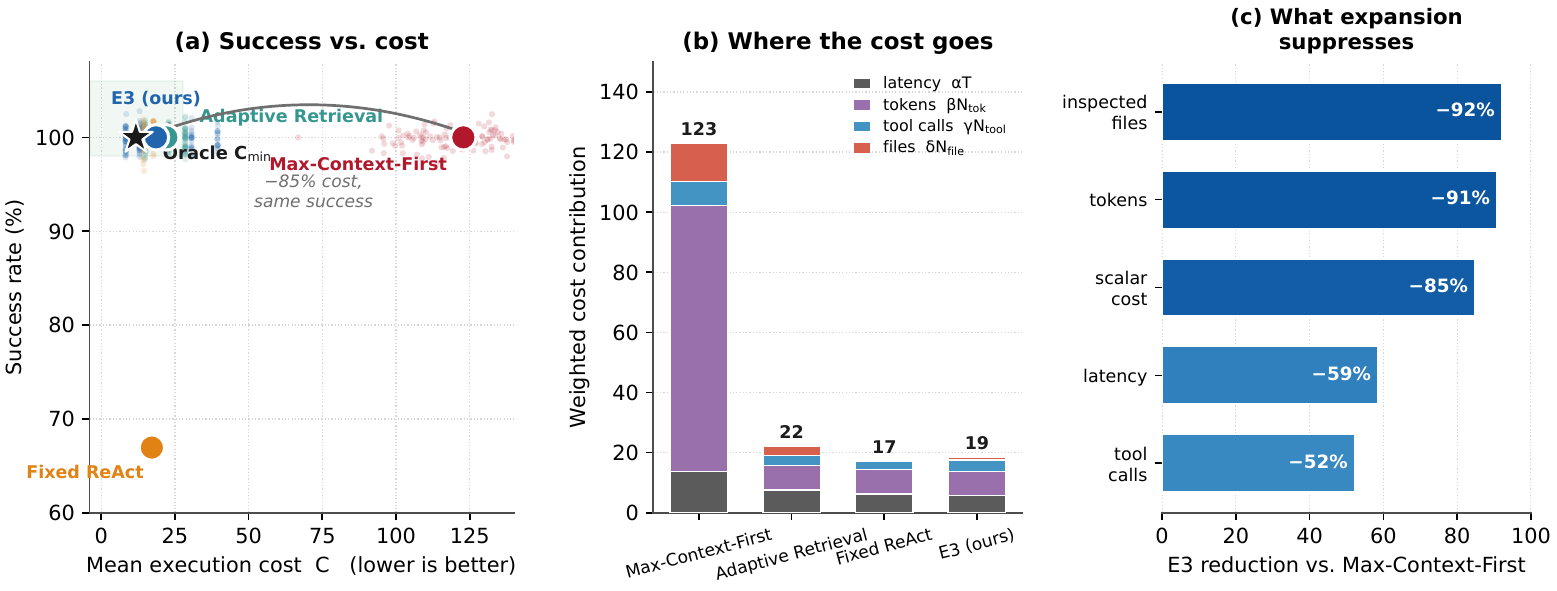}
\caption{(a) Success versus mean cost; faint clouds show the per-task runs
behind each policy mean. E3 lies nearest the oracle ($C_{\min}$); the strong
Adaptive Retrieval baseline is next; Max-Context-First is an order of magnitude
more expensive; Fixed ReAct is cheap but below the success frontier. (b) Weighted
cost breakdown: Max-Context-First's cost is dominated by tokens and fully
inspected files, the axes that progressive expansion avoids; Adaptive Retrieval is
far leaner but still pays to fully read every retrieved file. (c) E3's per-axis
reduction relative to Max-Context-First, concentrated on inspected files ($92\%$)
and tokens ($91\%$).}
\label{fig:pareto}
\end{figure}

Table~\ref{tab:main} summarizes performance over all $121$ tasks. MCF attains
$100\%$ success but at a mean cost of $122.9$, inspecting $8.5$ files and spending
$4421$ tokens per task on average---an ACRR of $12.9$, i.e.\ roughly thirteen
times the necessary effort. Fixed ReAct is far leaner but solves only $66.9\%$ of
tasks, failing every Level-3 task because it never traces the indirect site (its
ACRR of $1.29$ is computed only over the tasks it does solve). The strong
\emph{Adaptive Retrieval} baseline closes that reliability gap---$100\%$ success by
always tracing imports on cross-file tasks---at a mean cost of $22.1$ (ACRR
$1.21$), roughly one-fifth of MCF's: a genuinely competitive adaptive agent, not a
straw man. E3 matches the $100\%$ success of both at a mean cost of $18.6$ (ACRR
$0.55$). Relative to MCF, E3 reduces latency by $58.6\%$, tokens by $90.9\%$, tool
calls by $52.3\%$, inspected files by $92.2\%$, and scalar cost by $84.9\%$;
relative to the much tougher Adaptive Retrieval baseline it still cuts scalar cost
by $16.0\%$ and inspected files by $66.8\%$ at equal success, while---being willing
to expand on hard tasks---spending about $14\%$ \emph{more} tool calls, a trade we
examine by level next. Figure~\ref{fig:pareto}(a) places the policies on a
success--cost plane: E3 sits nearest the oracle, Adaptive Retrieval next, MCF far
to the right, and Fixed ReAct below the success frontier. Figure~\ref{fig:pareto}(b)
shows \emph{where} the baselines' cost goes---MCF overwhelmingly into tokens and
fully inspected files, the two axes that progressive expansion suppresses. Read as
a whole, the four policies trace the trade-off the paper targets: Fixed ReAct is
cheap but \emph{unreliable} ($66.9\%$ success), MCF and AR are reliable but
\emph{wasteful}, and only E3---an up-front scope judgment backed by verified
expansion---is at once the most reliable and the leanest fully-successful policy.
Efficiency here is not spending \emph{less} but spending \emph{correctly}: the same
judgment that strips waste from simple tasks also supplies the coverage a
cheap-but-blind policy lacks on hard ones, which is why E3 alone sits on both the
success and the cost frontier.

\begin{table}
\tbl{Main results on MSE-Bench (means over 121 tasks). Cost axes as in
Eq.~\eqref{eq:cost}; $C$ is the scalar cost; ACRR is averaged over successful runs
only (a cheap failure is not an efficiency), so Fixed ReAct's ACRR reflects only
the $81$ tasks it solves. Best cost/ACRR among the fully-successful ($100\%$)
policies in bold.}
{\setlength{\tabcolsep}{4pt}%
\begin{tabular}{@{}lrrrrrrr@{}} \toprule
Policy & Succ.\% & Lat. & Tok. & Tool & File & $C$ & ACRR \\ \midrule
Max-Context-First  & 100.0 & 13.80 & 4421 & 15.9 & 8.46 & 122.85 & 12.90 \\
Fixed ReAct        & 66.9  & 6.26  & 404  & 5.7  & 0.00 & 17.16  & 1.29 \\
Adaptive Retrieval & 100.0 & 7.56  & 410  & 6.6  & 1.99 & 22.08  & 1.21 \\
\textbf{E3 (ours)} & 100.0 & 5.71  & 403  & 7.6  & 0.66 & \textbf{18.55} & \textbf{0.55} \\ \midrule
Oracle ($C_{\min}$) & 100.0 & -- & -- & -- & -- & 11.74 & 0.00 \\ \bottomrule
\end{tabular}}
\label{tab:main}
\end{table}

\subsection{Redundancy grows worst on the simplest tasks}\label{subsec:bylevel}
\begin{figure}
\centering
\includegraphics[width=0.82\linewidth]{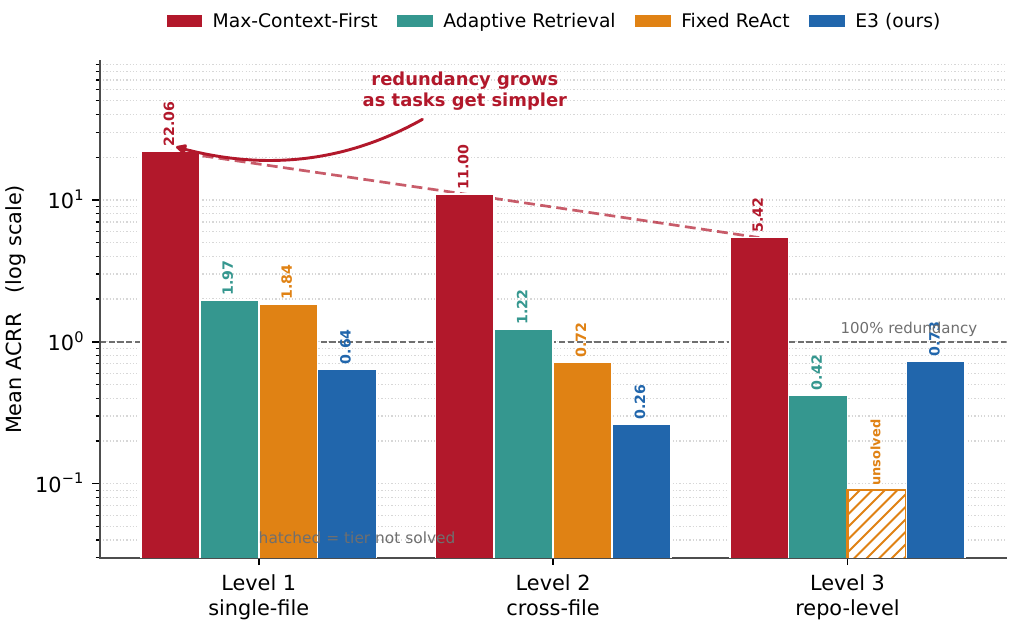}
\caption{Mean ACRR by task level (log scale), for all four policies.
Max-Context-First's redundancy is largest on the \emph{simplest} tasks. E3 stays
low and roughly flat while solving every tier; the strong Adaptive Retrieval
baseline also solves every tier and is competitive on repository-level tasks but
far costlier on simple ones; hatched bars indicate a tier the policy does not
solve (Fixed ReAct on Level~3).}
\label{fig:acrr}
\end{figure}

Figure~\ref{fig:acrr} reports ACRR by tier. The redundancy of MCF is \emph{highest
on the easiest tasks}: ACRR $22.1$ at Level~1, $11.0$ at Level~2, and $5.4$ at
Level~3. This is the quantitative signature of the motivating anecdote---the
simpler the task, the larger the fraction of effort wasted when the agent insists
on reading everything. We are candid that the \emph{monotonicity} is partly
mechanical: MCF's actual cost is nearly constant (it always reads the whole
repository), while $C_{\min}$ grows with the tier, so the ratio
$(C_{\text{act}}-C_{\min})/C_{\min}$ is necessarily largest where the denominator
is smallest, namely Level~1. We therefore present it as a \emph{descriptive
signature} of the fixed-overhead instinct rather than a discovered scaling law.
The practical consequence is nonetheless real: the absolute waste---files and
tokens read for nothing---is incurred on exactly the tasks that least warrant it.
E3 keeps ACRR low and roughly flat across tiers ($0.64$, $0.26$, $0.73$) and,
unlike Fixed ReAct, solves all Level-3 tasks (hatched bars mark unsolved tiers).
Fixed ReAct's low Level-3 ACRR is illusory: it is cheap because it \emph{fails}
(we now leave its unsolved tier blank rather than reporting the spurious negative
mean that averaging over failures previously produced). Notably, on Level~3 the
Adaptive Retrieval baseline (ACRR $0.42$) is \emph{leaner than E3} (ACRR $0.73$):
because E3 starts optimistically, it pays an extra expansion on the deceptive
tasks it under-scopes, whereas AR pays the full tracing cost up front. E3's
advantage is thus concentrated where the thesis says it should be---the simple
Level-1 and Level-2 tasks, where E3 is $45\%$ and $43\%$ cheaper than AR---while
on genuinely hard tasks a thorough adaptive agent is competitive. This is the
honest shape of the result, and it is precisely the shape the thesis predicts.

\subsection{The motivating case}\label{subsec:gmail}
Table~\ref{tab:gmail} isolates the Gmail-icon task ($C_{\min}=6.0$). By
construction the over-cautious MCF policy inspects all seven project files, spends
$1844$ tokens, and reaches an ACRR of $10.1$---over $1000\%$ redundancy on a
two-line edit. We are careful about this figure: it is the redundancy of the
\emph{simulated} over-cautious policy, designed to reproduce the \emph{order of
magnitude} of the lived experience that opened the paper, not a measurement of any
particular frontier agent. Even the strong Adaptive Retrieval baseline, which
reads only the single retrieved file, still spends $19.6$ (ACRR $2.27$) because it
fully loads that file and runs the heavy check. E3 inspects \emph{zero} irrelevant
files: its estimator takes the localized single-file fast path, and it edits after
a single localization step, for an ACRR of $0.59$---an $85.8\%$ cost reduction
versus MCF and $51\%$ versus Adaptive Retrieval. Figure~\ref{fig:anatomy}(a)
unrolls this same case action by action: MCF's cumulative cost climbs relentlessly
as it inspects every project file, whereas E3 reaches a verified edit in a handful
of steps close to the oracle floor.

\begin{table}
\tbl{The motivating Gmail-icon task ($C_{\min}=6.0$). E3 pulls no irrelevant
file into context; even the adaptive baseline pays to fully read the one file it
retrieves.}
{\begin{tabular}{@{}lrrrr@{}} \toprule
Policy & Files & Tokens & $C$ & ACRR \\ \midrule
Max-Context-First  & 7 & 1844 & 66.78 & 10.13 \\
Fixed ReAct        & 0 & 420  & 15.80 & 1.63 \\
Adaptive Retrieval & 1 & 531  & 19.60 & 2.27 \\
\textbf{E3 (ours)} & 0 & 248  & \textbf{9.51} & \textbf{0.59} \\ \bottomrule
\end{tabular}}
\label{tab:gmail}
\end{table}

\subsection{Ablation: estimation cuts cost, expansion preserves success}\label{subsec:ablation}
\begin{figure}
\centering
\includegraphics[width=\linewidth]{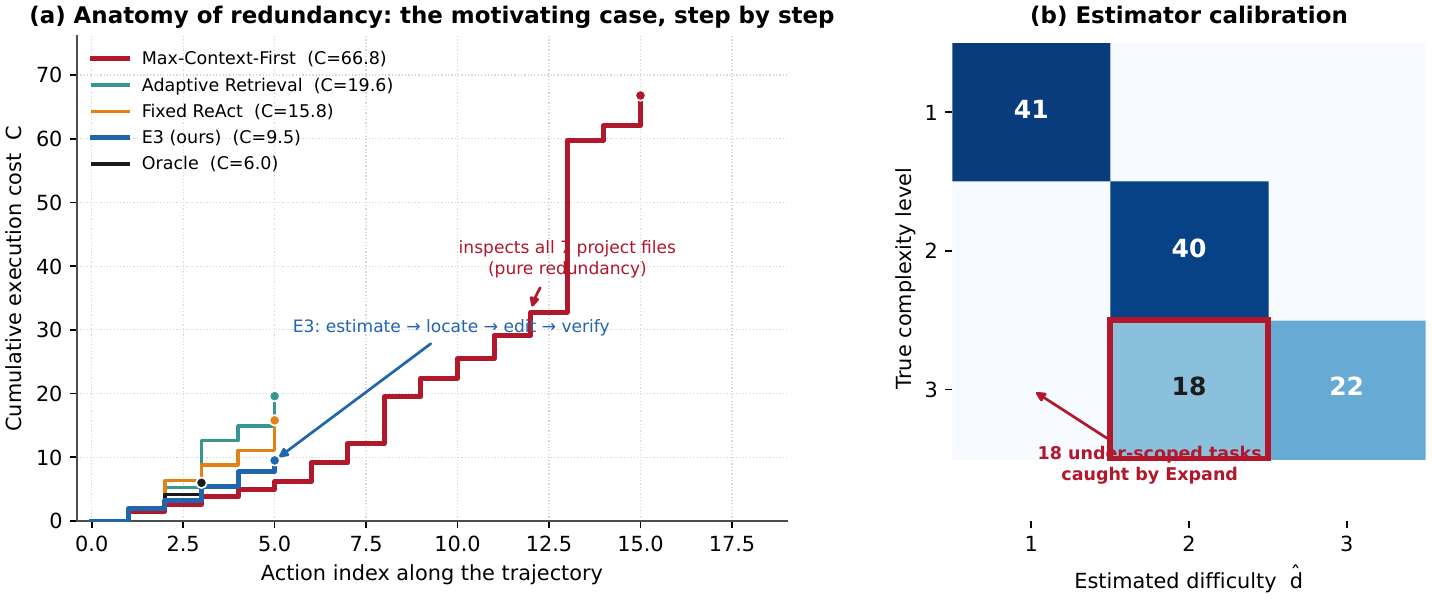}
\caption{Anatomy of redundancy and the safety net. (a) The motivating Gmail-icon
task unrolled as a running-cost trajectory: Max-Context-First's cost climbs step
by step as it inspects all seven project files, the Adaptive Retrieval baseline
still pays to fully read the one retrieved file and run the heavy check, while E3
reaches a verified edit in a handful of actions near the oracle floor. (b)
Estimator calibration over all $121$ tasks (rows: true complexity level; columns:
estimated difficulty $\hat d$): the transparent estimator is accurate on
Levels~1--2 and on \emph{obvious} Level-3 tasks, and deliberately under-scopes the
$18$ \emph{deceptive} Level-3 tasks (highlighted; predicted difficulty~$2$ rather
than $3$)---exactly the cases the Expand stage recovers, which is why optimistic
estimation is safe.}
\label{fig:anatomy}
\end{figure}

Table~\ref{tab:ablation} removes one E3 stage at a time. Without \emph{Expand}, the
agent cannot recover under-estimated tasks: success falls to $85.1\%$ (it loses
all $18$ deceptive Level-3 tasks) even though cost drops, confirming that
expansion is the safety net that makes optimistic estimation safe. Without
\emph{Estimate}, the agent always starts minimal and pays for expansion on every
non-trivial task: success stays at $100\%$ but cost rises by $20\%$ overall and by
$36\%$ on Level~3 ($47.0$ versus $34.6$). The two stages are complementary:
estimation reduces cost, and expansion protects reliability.
Figure~\ref{fig:anatomy}(b) shows why optimistic estimation is safe: the
transparent estimator under-scopes precisely the $18$ deceptive Level-3 tasks,
predicting difficulty~$2$ rather than $3$ (their localized wording exposes only
the two direct hits, so the estimator stops one dependency trace short), and these
are exactly the runs that Expand recovers.

\begin{table}
\tbl{Ablation over E3 stages (means). Removing Expand drops success; removing
Estimate raises cost, especially on Level~3.}
{\begin{tabular}{@{}lrrrr@{}} \toprule
Variant & Succ.\% & $C$ & ACRR & L3 $C$ \\ \midrule
E3 (full)        & 100.0 & 18.55 & 0.55 & 34.59 \\
\;\; $-$ Expand  & 85.1  & \textbf{14.88} & 0.47 & 23.48 \\
\;\; $-$ Estimate& 100.0 & 22.21 & 0.71 & 47.01 \\ \bottomrule
\end{tabular}}
\label{tab:ablation}
\end{table}

\subsection{Robustness I: sensitivity to the cost weighting}\label{subsec:sensitivity}
A natural worry is that the ordering depends on our chosen weights
$(\alpha,\beta,\gamma,\delta)$---in particular on the deliberately large file
weight $\delta$, which rewards exactly the axis E3 spares. Because the scalar cost
is linear in the four axes, a policy's mean cost under \emph{any} weight vector is
that vector dotted with its mean per-axis usage, so we can evaluate the ordering
over a large sample of weightings without re-simulating. We draw $4000$ weight
vectors from broad, deliberately unfavourable ranges---$\alpha\!\in\![0.5,2]$,
$\beta\!\in\![0,0.1]$, $\gamma\!\in\![0,2]$, $\delta\!\in\![0,3]$, so a draw may
switch the token or file axis off entirely---and ask how often E3 remains the
cheapest policy that still solves every task. It does so in $99.8\%$ of weightings.
Against MCF, E3 is cheaper in $100\%$ of draws (median reduction $86.6\%$; $5$th
percentile $74.0\%$). Against the far tougher Adaptive Retrieval baseline, E3 is
cheaper in $99.8\%$ of draws (median $9.3\%$; interdecile range $3.5$--$21.2\%$),
and \emph{even when the file penalty is removed entirely} ($\delta{=}0$)---the
weighting most hostile to E3's advantage---it remains cheaper than AR in $96.7\%$
of draws and cheaper than MCF in $100\%$. Figure~\ref{fig:robust}(b) plots the two
reduction distributions. The conclusion does not hinge on the cost model: it holds
because E3 simply issues fewer and lighter actions, and because ACRR is in any
case normalized by each task's own $C_{\min}$ and hence scale-free in the overall
magnitude of the weights.

\subsection{Robustness II: held-out instruction wording}\label{subsec:robustness}
The sharpest objection to the estimator is that its lexical rules
(Section~\ref{subsec:estimate}) were written against the very templates that
generate the benchmark instructions, so its accuracy could be circular. We test
this directly. We build a \emph{held-out} copy of MSE-Bench in which every
instruction is paraphrased into wording whose vocabulary is \emph{disjoint} from
the estimator's keyword lists and which removes the file-plus-quoted-literal fast
path---for instance the obvious Level-3 template ``refactor \dots\ across the
entire codebase \dots\ the re-exported settings key'' becomes ``give \dots\ a new
name project-wide \dots\ the settings alias that other modules import''. Nothing
else changes: identical levels, sites, repositories, and oracles.

\begin{figure}
\centering
\includegraphics[width=\linewidth]{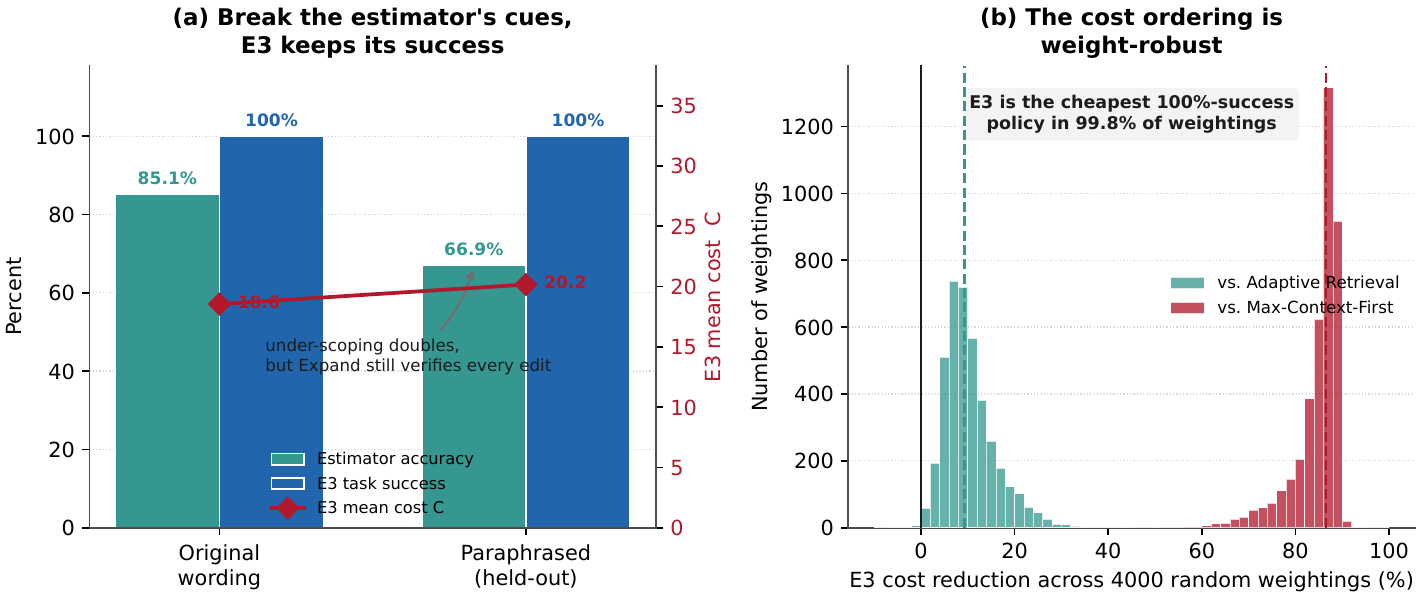}
\caption{Robustness of the claims. (a) Held-out instruction wording that breaks
the estimator's lexical cues nearly halves its accuracy and doubles under-scoping,
yet E3's success stays at $100\%$ and its mean cost rises only $8.7\%$ because
Expand verifies and recovers every under-scoped task. (b) Distribution of E3's
cost reduction over $4000$ random cost weightings: E3 is cheaper than
Max-Context-First in every draw and cheaper than the strong Adaptive Retrieval
baseline in $99.8\%$ of them (dashed lines mark medians); E3 is the cheapest
fully-successful policy in $99.8\%$ of weightings.}
\label{fig:robust}
\end{figure}

Under this held-out wording the estimator can no longer take its fast paths, and
its exact-level accuracy drops from $85.1\%$ to $66.9\%$; the fraction of tasks it
\emph{under-scopes} more than doubles, from $14.9\%$ to $33.1\%$ (it now
under-scopes \emph{all} $40$ Level-3 tasks, obvious as well as deceptive). Yet
E3's success stays at $100.0\%$, and its mean cost rises only from $18.55$ to
$20.17$---an $8.7\%$ increase, still $84\%$ below MCF and $9\%$ below Adaptive
Retrieval---because progressive expansion recovers every newly under-scoped task
(Figure~\ref{fig:robust}(a)). This is the heart of our answer to the circularity
concern: the headline efficiency is a property of the Estimate--Expand
\emph{architecture}---an optimistic start backed by verified expansion---not of the
estimator's keywords matching the benchmark. A better estimator would shift cost
down by expanding less often; a worse one (as here) shifts it up; neither changes
the success guarantee that Expand provides.

\subsection{Real-LLM validation: the LLM-Case harness}\label{subsec:llmcase}

The capability-controlled simulator is what lets us isolate trajectory shape, but
it invites one sharp objection: \emph{every component is synthetic}---the policies
are hand-written functions, the benchmark and oracle are self-authored, and the
head-line reduction is measured against a mechanical worst case. To confront this
directly we implement and release \textbf{LLM-Case}, a companion harness that
re-poses the \emph{same} question---does an agent over-gather context on simple
tasks, and does Estimate--Execute--Expand remove that redundancy without hurting
success?---with a real model in the loop, real tools on real code, real tests, and
a \emph{measured} oracle. It removes each synthetic crutch in turn.

\paragraph{No hand-written policies.} The three policies are nothing but
\emph{system prompts}---\textsc{mcf-thorough} (``read everything first''),
\textsc{react}, and \textsc{e3}---handed to the \emph{same} model over the
\emph{same} six tools (\texttt{list\_dir}, \texttt{grep}, \texttt{read\_file},
\texttt{edit\_file}, \texttt{run\_tests}, \texttt{finish}). The model chooses every
action through native tool calls, so swapping the prompt is the only difference
between the over-cautious default, a plain ReAct loop, and E3; the baseline is thus
a genuine agent following a realistic ``be thorough'' instruction, not a mechanism
that reads every file by construction.

\paragraph{Real code, real coupling.} Tasks edit a genuine vendored open-source
library---the MIT-licensed \texttt{toml} 0.10.2 package\footnote{Vendored
unmodified from \texttt{https://github.com/uiri/toml}; see the accompanying code
repository's \texttt{NOTICE} and \texttt{LICENSE} files.}---whose seven Python
modules (ten files in all) provide real multi-file structure.
Table~\ref{tab:llmtasks} lists the five tasks across the same three tiers as
MSE-Bench, but here each tier is a property of the library's real import graph. The
Level-3 instruction (``rename the decoder base class'') \emph{sounds} local, yet a
correct change must also reach \texttt{toml/ordered.py}, which imports and
subclasses \texttt{TomlDecoder}---genuine coupling that punishes under-scoping and
exercises E3's \emph{Expand} stage.

\paragraph{A measured oracle, not an asserted one.} Success is decided by
\emph{actually running \texttt{pytest}} against hidden per-task acceptance graders,
and $C_{\min}$ is \emph{measured} by running each gold patch through the same
instrumented tools rather than asserted (Table~\ref{tab:llmoracle}). Two facts
validate the design without invoking any model. First, the harness is
self-consistent: every gold patch passes its hidden grader, the pristine repository
\emph{fails} every grader (so the graders discriminate rather than being vacuously
green), and the measured oracle inspects \emph{exactly} the one, two, or three
files each patch must touch---never the remaining seven-to-nine files of the
repository. Second, the measured token floor is set by \emph{which} file must be
read, not how many: renaming a class inside the large \texttt{decoder.py} costs an
order of magnitude more unavoidable context ($\sim\!10{,}000$ tokens, tasks
\texttt{L2\_commentdecoder} and \texttt{L3\_decoderbase}) than the two-line
Level-1 edits ($\sim\!210$ tokens), so a policy that pulls all ten files into
context pays that floor several times over on tasks that truly need one.

\begin{figure}
\centering
\includegraphics[width=\linewidth]{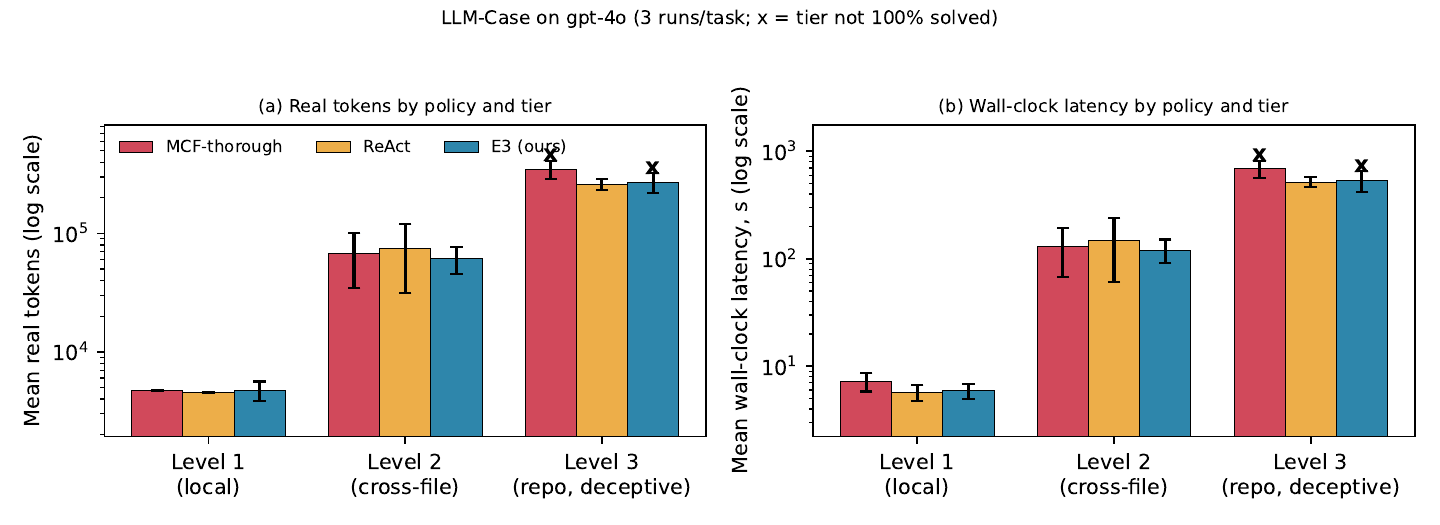}
\caption{LLM-Case on \textsc{gpt-4o}, by policy and task tier (three runs per
task; log scales; error bars are $\pm 1$ standard deviation across a tier's runs;
``x'' marks a tier a policy did not solve on every run). \textbf{(a)} Mean real
prompt-plus-completion tokens; \textbf{(b)} mean wall-clock latency (seconds). On
the simple Level-1/2 tiers the policies are close and all succeed; the deceptive
Level-3 refactor is an order of magnitude costlier---in both tokens and
time---for \emph{every} policy, and the heaviest-reading trajectories (the
thorough agent, and one throttled E3 run) are the ones that fail. E3 is the
leanest and fastest policy overall.}
\label{fig:llmtokens}
\end{figure}

\paragraph{A non-straw-man head-line.} The harness reports E3's reduction against
\emph{two} real agents---a plain ReAct agent and the ``be thorough'' agent---at
matched success, plus ACRR against the measured oracle of
Table~\ref{tab:llmoracle}. Here the token axis is the provider's real
prompt-plus-completion count, files are the distinct files the model actually
pulled into context, and success is a real test pass, so none of the four
synthetic crutches survives. We release LLM-Case so this real-model comparison is a
turnkey run---a single command against any Anthropic, OpenAI, or
OpenAI-compatible endpoint---rather than a bespoke re-implementation. Running it
\emph{three times per cell} on \textsc{gpt-4o} over the five tasks
(Table~\ref{tab:llmresults}, Figure~\ref{fig:llmtokens}) confirms that the
phenomenon survives contact with a real model, but in a \emph{milder and more
nuanced} form than the capability-controlled simulation, and we report it as such.
First, all three prompted policies are far more frugal than the mechanical MCF of
Section~\ref{sec:setup}: \textsc{gpt-4o} inspects only one to four files even under
the explicit ``read everything first'' instruction, so the gross file-over-reading
the simulator models (eight-plus files) does not reproduce in a frontier
model---the redundancy is real but bounded, exactly as our \emph{upper-bound}
framing of MCF anticipated. Second, E3 is the \emph{leanest policy overall}---and,
on the axis a user actually feels, the \emph{fastest}: it spends $18\%$ fewer real
tokens than the thorough agent and $4\%$ fewer than ReAct, and its mean wall-clock
latency is likewise $18\%$ and $5\%$ lower (Table~\ref{tab:llmresults}), at
effectively equal task success. Against ReAct the time saving ($5\%$) even slightly
exceeds the token saving ($4\%$)---reading less context shortens each model
turn---while against the thorough agent both are near $18\%$; we nonetheless treat
latency as provider- and load-dependent and, on the throttled Level-3 tier,
confounded by rate limiting. Third, and most honestly,
this aggregate edge is \emph{small and non-uniform}. On the trivial Level-1 edits
the three policies are within noise of one another ($\sim\!4.5$--$5.4$k tokens):
E3's explicit estimate step is roughly \emph{cost-neutral}, not the overhead a
single unlucky run had earlier suggested, and not a win either. The clearest E3
gain is on the token-heavy \texttt{L2\_commentdecoder} rename ($76$k tokens versus
$100$k for the thorough agent and $116$k for ReAct), where reading the large
\texttt{decoder.py} once rather than repeatedly matters; on the smaller
\texttt{L2\_arrayseparator} E3 is actually the priciest. The deceptive Level-3
refactor is expensive for \emph{every} policy ($\sim\!260$--$350$k tokens) and,
revealingly, its reliability is \emph{stochastic}: this time the thorough
over-reading agent failed all three runs---twice by exhausting its step budget or
emitting a wrong edit, once by hitting the provider's tokens-per-minute rate limit
at $315$k cumulative tokens---whereas ReAct succeeded all three and E3 succeeded
both of its non-throttled runs. (In the earlier single-run pass the failure fell
instead on ReAct, which missed the hidden \texttt{ordered.py} coupling; that the
identity of the failing policy \emph{moves} across runs is itself the honest
finding.) Two robust conclusions survive this noise: a frontier model does not
grossly over-read, and \emph{aggressive over-reading carries a real operational
tax}---the highest-token trajectories are also the \emph{slowest} and the ones that
hit the step and rate limits and fail (the thorough agent's Level-3 runs ran
$575$--$863$\,s and none passed, versus ${\sim}450$\,s for E3's successful
runs)---so the leanest reliable policy, here E3, is preferable even when a capable
model makes the raw edit easy. The real-model data thus relocates
E3's benefit from ``always dramatically cheaper'' (the simulation's headline) to
``the leanest policy overall, and one that does not spend itself into failure as
hidden coupling grows.'' These are three trajectories per cell on one model under a
low provider rate limit, reported as a \emph{case study} rather than a powered
benchmark; LLM-Case's five objective, cheaply gradable edits isolate \emph{scope
estimation} from raw capability, and the identical agent loop points at additional
models and at SWE-bench-style tasks~\citep{Jimenez2024SWEbench} unchanged.

\begin{table}
\tbl{LLM-Case tasks over the real \texttt{toml} 0.10.2 package. Each complexity
tier is a property of the library's import graph; the Level-3 instruction sounds
local but a correct patch must also update \texttt{toml/ordered.py}, which
subclasses the renamed base class.}
{\setlength{\tabcolsep}{5pt}%
\begin{tabular}{@{}lllr@{}} \toprule
Task & Level & Change & Files \\ \midrule
\texttt{L1\_version}        & 1 (local)           & bump the package version        & 1 \\
\texttt{L1\_spec}           & 1 (local)           & bump the supported-spec version & 1 \\
\texttt{L2\_commentdecoder} & 2 (cross-file)      & rename a public decoder class   & 2 \\
\texttt{L2\_arrayseparator} & 2 (cross-file)      & rename a public encoder class   & 2 \\
\texttt{L3\_decoderbase}    & 3 (repo, deceptive) & rename the decoder base class   & 3 \\ \bottomrule
\end{tabular}}
\label{tab:llmtasks}
\end{table}

\begin{table}
\tbl{Measured oracle for LLM-Case: the minimum-sufficient trajectory run through
the real tools. Tokens are the real byte-size floor of the gold files; $C_{\min}$
is the scalar cost of Eq.~\eqref{eq:cost} over the deterministic token,
tool-call, and file axes; the measured wall-clock latency (of order a second per
task and machine-dependent) is omitted here so that $C_{\min}$ stays
deterministic, though the released data retains it. The floor is governed by file
\emph{size}, not count---the large \texttt{decoder.py} dominates
\texttt{L2\_commentdecoder} and \texttt{L3\_decoderbase}.}
{\setlength{\tabcolsep}{6pt}%
\begin{tabular}{@{}lrrrr@{}} \toprule
Task & Files & Tool calls & Tokens & $C_{\min}$ \\ \midrule
\texttt{L1\_version}        & 1 & 4 & 212   & 7.7 \\
\texttt{L1\_spec}           & 1 & 4 & 214   & 7.8 \\
\texttt{L2\_commentdecoder} & 2 & 6 & 9975  & 205.5 \\
\texttt{L2\_arrayseparator} & 2 & 6 & 2712  & 60.2 \\
\texttt{L3\_decoderbase}    & 3 & 8 & 10065 & 209.8 \\ \bottomrule
\end{tabular}}
\label{tab:llmoracle}
\end{table}

\begin{table}
\tbl{LLM-Case on \textsc{gpt-4o}, \emph{three} autonomous runs per cell: mean
distinct files inspected and mean \emph{real} prompt-plus-completion tokens per
policy, and per-policy success over all $15$ runs. A frontier model stays
frugal---one to four files, never the eight-plus the simulator's MCF
assumes---even under a ``read everything first'' prompt. On the clean Level-1/2
cells (every policy $3/3$) the three policies are close: E3 wins the token-heavy
\texttt{L2\_commentdecoder} but is within noise on the trivial edits and priciest
on \texttt{L2\_arrayseparator}, so the estimate step is roughly cost-neutral rather
than a uniform win. Level-3 is costly for all and its reliability is stochastic:
two runs terminated on the provider's tokens-per-minute limit (HTTP~429) rather
than task failure ($^\dagger$; MCF and E3 once each), and excluding those the
thorough over-reader still failed every Level-3 run ($^\ddagger$; step-budget
exhaustion or a wrong edit) while ReAct and E3 solved every non-throttled run. E3
is the leanest policy overall; the bottom \emph{Mean lat.}\ row gives mean
wall-clock latency per policy (seconds), on which E3 is also the fastest, though
latency is provider- and load-dependent and the Level-3 runs include rate-limited
ones. Bold marks the leanest token count per row and the leanest mean latency;
reported as a case study, not a powered benchmark.}
{\setlength{\tabcolsep}{4pt}%
\begin{tabular}{@{}lrrrrrr@{}} \toprule
 & \multicolumn{2}{c}{MCF-thorough} & \multicolumn{2}{c}{ReAct} & \multicolumn{2}{c}{E3 (ours)} \\
Task & File & Tok. & File & Tok. & File & Tok. \\ \midrule
\texttt{L1\_version}        & 1.0 & 4708  & 1.0 & 4542           & 0.3 & \textbf{4121} \\
\texttt{L1\_spec}           & 1.0 & 4682  & 1.0 & \textbf{4577}  & 1.0 & 5356 \\
\texttt{L2\_commentdecoder} & 2.0 & 99555 & 2.7 & 116406         & 2.0 & \textbf{76058} \\
\texttt{L2\_arrayseparator} & 2.3 & 35228 & 2.3 & \textbf{34000} & 2.0 & 45991 \\
\texttt{L3\_decoderbase}    & 3.7 & 348884$^{\dagger\ddagger}$ & 3.0 & \textbf{259866} & 3.0 & 270990$^{\dagger}$ \\ \midrule
Mean                        & 2.0 & 98611 & 2.0 & 83878          & 1.7 & \textbf{80503} \\
Mean lat.\ (s)              & \multicolumn{2}{c}{192.3} & \multicolumn{2}{c}{165.9} & \multicolumn{2}{c}{\textbf{157.6}} \\
Success                     & \multicolumn{2}{c}{$80\%^{\dagger\ddagger}$} & \multicolumn{2}{c}{$100\%$} & \multicolumn{2}{c}{$93\%^{\dagger}$} \\ \bottomrule
\end{tabular}}
\label{tab:llmresults}
\end{table}

\section{Power-System Engineering Case Study}\label{sec:casestudy}

Our framing borrows directly from power-system numerics, and the analogy is more
than rhetorical. We solve the AC power flow of a three-bus system (one slack, two
PQ buses) with polar Newton--Raphson~\citep{TinneyHart1967}, and measure how the
number of iterations to convergence---and whether the solver converges at
all---depends on the quality of the initial operating point $x_0$.

\begin{figure}
\centering
\includegraphics[width=\linewidth]{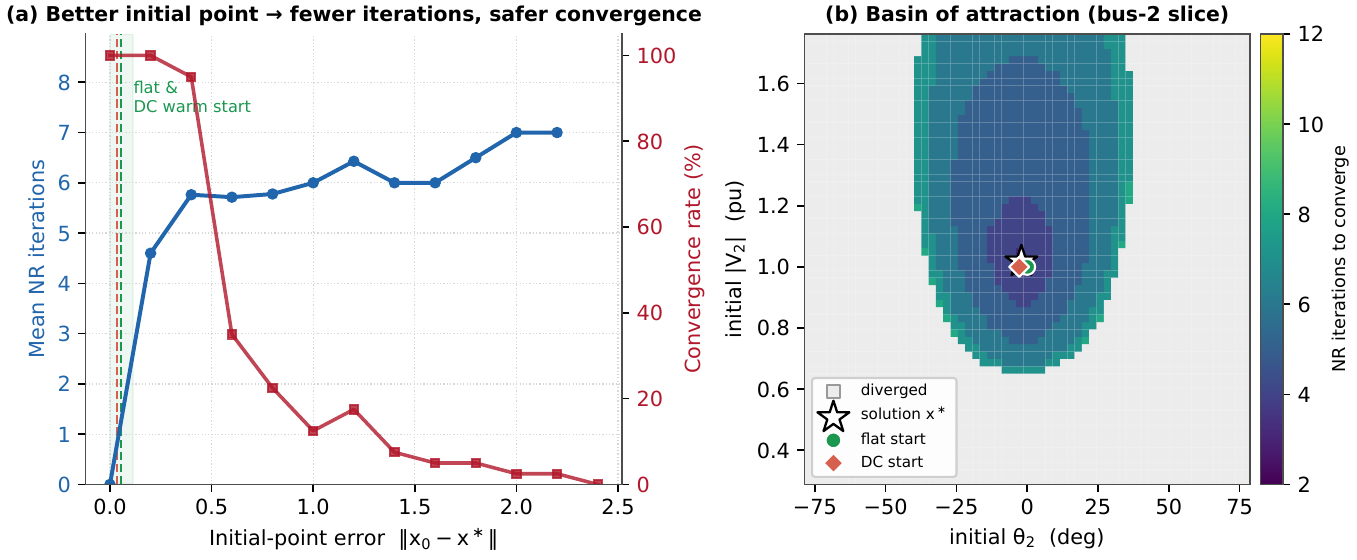}
\caption{Power-flow analogy on a three-bus system. (a) A better initial operating
point (smaller $\lVert x_0-x^\star\rVert$) yields fewer Newton--Raphson iterations
(blue) and far higher convergence reliability (red); the flat and DC warm starts
sit in the low-error, reliable regime. (b) The basin of attraction over a
two-dimensional slice of the initial guess for bus~2 (bus~1 held at its
solution): points near the solution converge in a few iterations (dark), while
distant guesses diverge (grey). The cheap warm starts lie deep inside the
basin---the numerical counterpart of E3's task-state estimate landing in a
workable region rather than searching an unstructured space.}
\label{fig:pf}
\end{figure}

Figure~\ref{fig:pf}(a) reports iterations and convergence rate as a function of the
initial-point error $\lVert x_0 - x^\star\rVert$. A \emph{flat start} and a
\emph{DC-estimated warm start}---both cheap, structured guesses near the
solution---converge in three iterations with $100\%$ reliability. As the initial
point moves away from the solution, iteration count rises and, more importantly,
convergence collapses: reliability falls to $35\%$ at error $0.6$ and to near zero
beyond error $\sim\!2.4$, where Newton--Raphson diverges. Figure~\ref{fig:pf}(b)
makes the same point geometrically: the basin of attraction---the set of initial
points from which Newton--Raphson converges---is a compact region around the
solution, and the cheap warm starts lie deep inside it while distant guesses fall
into the surrounding divergent region. The parallel to agents is
\emph{structural} rather than exact: a good initial operating point is not the
answer, but it makes the subsequent refinement short and stable, whereas a poor one
causes a long, unstable search that often fails outright---the solver analogue of
an agent wandering a repository without a scope estimate. We offer the analogy as
an engineering intuition that motivated E3's design, not as evidence for its
agent-side results, which stand or fall on MSE-Bench.

This connection suggests a broader principle we term \emph{engineering-grounded
efficiency}---an efficiency-oriented instance of \emph{engineering-grounded AI}
(EGAI), the view that an agent's reasoning and action should be anchored in the
physical, model-based, and procedural reality of an engineered task rather than in
unconstrained search~\citep{Yin2026EGAI}. A competent agent, like a competent
solver, should not search an unstructured state space exhaustively. It should
exploit task structure,
observable environment state, and verifiable checks to anchor itself in a
reasonable initial operating region, and expand computation only when the evidence
demands it. Estimation of $x_0$ and progressive expansion are the agentic
realizations of warm-starting and controlled iteration. The same
warm-start-and-refine discipline recurs across power-system optimization and
analysis more broadly---from integrated-energy-system operation and
multi-energy-flow scheduling to distribution-network assessment and
converter-dominated grid studies---where a structured initial point keeps large,
nonconvex problems fast and
stable~\citep{Yin2021IES,Zhang2022ICPET,Xu2022TCES,Lu2022CCHP,Jiang2020SGEE,Kong2023Energies}.

\section{Discussion}\label{sec:discussion}

\paragraph{When \emph{not} to think deeply.} The field has invested heavily in
deeper reasoning, longer context, and richer tool use. Our results argue for a
complementary and under-studied competence: recognizing when a task is simple and
declining to deliberate. On the easiest tasks, the cost of misjudging complexity
is largest in relative terms (Figure~\ref{fig:acrr}); this is precisely where a
cheap, confident estimate pays off most.

\paragraph{Relation to routing and adaptive effort.} E3 is compatible with, and
complementary to, per-step effort selection~\citep{Yang2026Ares}, LLM-versus-agent
routing~\citep{Wang2026BoundaryRouter}, and paradigm
routing~\citep{Zhou2026SelectThenSolve}. Those methods adjust \emph{how much to
think} or select \emph{which engine to use} from a fixed set; E3 instead performs
\emph{execution-scope estimation}---an anticipatory judgment of \emph{what to
understand before acting} and \emph{how little context suffices}---made safe by
verifying and expanding (Table~\ref{tab:paradigms}). The two are orthogonal and
composable: a natural integration uses an effort router \emph{inside} each E3
stage, once the scope estimate has fixed \emph{what} the stage must accomplish.

\paragraph{Limitations and threats to validity.} The most important limitation is
that we evaluate \emph{policies in a capability-controlled simulator}, not a
specific deployed LLM agent: no language model is invoked anywhere in our
experiments. This is deliberate---it is what lets us construct an exact oracle and
attribute every difference to trajectory shape rather than to a model sample---but
it means our numbers are properties of a cost model. To confront this head-on we
implement and release \emph{LLM-Case} (Section~\ref{subsec:llmcase}), a real-model
harness that removes the synthetic-pipeline objections in apparatus---a real LLM
choosing every action, real vendored code, real \texttt{pytest} grading, and a
measured oracle---and whose internal self-consistency we verify here; its
head-line comparison we populate here with a case-study run on \textsc{gpt-4o}
(three runs per task; Section~\ref{subsec:llmcase}), which finds the phenomenon
survives in a milder form---a frontier model is already frugal, yet E3 is the
leanest and fastest policy overall, and on the deceptive repository-level task---where success
turns stochastic and the heaviest-reading trajectories are the ones that fail,
some even throttled by the provider's rate limit---E3 stays among the lean and
reliable runs. Scaling this to more models and to SWE-bench-style tasks is the
natural next step. We have tried to make
the simulation earn trust rather than assume it. Three objections a reader might
raise, and our responses. \emph{(i) A straw-man baseline.} MCF is an explicit
upper-bound stress model, so we also compare against a strong, genuinely adaptive
retrieval-augmented baseline that solves every task; E3 still wins on overall cost,
and is bettered only on the hardest tier
(Sections~\ref{subsec:main}--\ref{subsec:bylevel}). \emph{(ii) A circular
estimator.} Because the estimator's rules could be tuned to the templates, we
re-run everything on held-out paraphrased wording that breaks those rules;
accuracy falls sharply, but E3's success is unchanged and its cost barely moves
(Section~\ref{subsec:robustness}). \emph{(iii) A convenient cost model.} We sweep
$4000$ weightings, including ones that zero out the very axes E3 favours, and the
ordering survives (Section~\ref{subsec:sensitivity}). What the simulator still
cannot tell us is how a real model's \emph{edit accuracy} interacts with scope: a
model might make errors that only broader context would prevent, raising the
optimal operating point and shifting the estimate/expand balance; capturing that
requires running the LLM-Case harness of Section~\ref{subsec:llmcase} against a
provider at scale. Our estimator is likewise a transparent lexical rule;
a learned or LLM-backed estimator may generalize better and mis-estimate
differently. Finally, MSE-Bench's indirect-dependency mechanism captures one
important source of hidden complexity (aliases and re-exports) but not all
(dynamic dispatch, configuration coupling, runtime reflection). We view MSE-Bench
as a controlled complement to capability benchmarks such as
SWE-bench~\citep{Jimenez2024SWEbench}, not a replacement.

\section{Conclusion and Future Work}\label{sec:conclusion}

We asked whether AI agents know when a task is simple, and found---in a
capability-controlled simulator---that a common default (gather maximal context,
then act) turns trivial edits into disproportionately expensive audits, with
redundancy \emph{largest on the simplest tasks}. We formalized minimum-sufficient
execution and the Agent Cognitive Redundancy Ratio, proposed the E3 framework
(Estimate, Execute, Expand) built around an estimated initial operating point and
progressive context expansion, and showed on MSE-Bench that E3 matches the
strongest baseline's success while cutting cost by $85\%$ and inspected files by
$92\%$---and that it also improves on a strong \emph{adaptive} baseline, retains
its success under held-out wording that breaks the estimator's cues, and stays the
cheapest fully-successful policy under essentially any cost weighting. A power-flow
case study supplied the engineering intuition behind the initial-operating-point
analogy.

To that end we implement and release \emph{LLM-Case}
(Section~\ref{subsec:llmcase})---three policies expressed purely as system prompts
over the same real tools, tasks on a genuine vendored library, success graded by
running \texttt{pytest}, and a measured oracle---and run it as a case study on
\textsc{gpt-4o}. The real model is markedly more frugal than the simulated worst
case, yet E3 is the leanest and fastest policy overall, and on the deceptive repository-level
task---where reliability turns stochastic and the heaviest-reading trajectories
are the ones that fail, some even throttled by the provider's rate limit---E3
stays among the lean and reliable runs. Scaling this comparison to more models and
to SWE-bench-style tasks, and placing an LLM-backed estimator behind the same
interface, is the natural next step. Further
work includes calibrated,
learned task-state estimators; extending MSE-Bench with additional
hidden-complexity mechanisms; and integrating per-step effort routers within E3
stages. We release our framework and benchmark to encourage work on an efficiency
that engineers already practice: knowing, quickly, when a problem is easy. More
broadly, we view task-aware execution as one facet of \emph{engineering-grounded AI}
(EGAI)---agents whose effort is matched to, and validated against, the engineering
reality of the task~\citep{Yin2026EGAI}.


\section*{Author contributions}
\textbf{Junjie Yin:} Conceptualization, Methodology, Software, Formal analysis,
Writing -- original draft. \textbf{Xinyu Feng:} Methodology, Validation,
Writing -- review \& editing.

\section*{Acknowledgements}
The authors thank CURENT engineering research center, Microsoft Research.

\section*{Disclosure statement}
The authors declare that they have no known competing financial interests or personal relationships that could have appeared to influence the work reported in this paper.

\section*{Funding}
This research received no specific grant.

\section*{Data availability statement}
The data and code that support the findings of this study are openly available in
the GitHub repository \emph{Do AI Agents Know When a Task Is Simple? Toward
Complexity-Aware Reasoning and Execution} at
\url{https://github.com/eejyin/Do-AI-Agents-Know-When-a-Task-Is-Simple-Toward-Complexity-Aware-Reasoning-and-Execution}.
The repository provides the reference implementation of E3, the MSE-Bench
benchmark, and the LLM-Case real-model validation harness---including the vendored
MIT-licensed \texttt{toml} tasks, the hidden acceptance graders, and the measured
oracle of Table~\ref{tab:llmoracle}---together with the power-flow case study and
the scripts that regenerate every table and figure in this paper. All simulated
results are deterministic given the reported random seed ($20260712$); the
LLM-Case token and latency measurements depend on the model provider and are
therefore reported per run.




\bibliographystyle{apacite}
\bibliography{cas-refs}

\end{document}